\documentclass[11pt,a4paper]{article}

\usepackage[hyperref]{naaclhlt2019}

\hypersetup{breaklinks=true}

\usepackage{times}
\usepackage{latexsym}

\usepackage{graphicx}
\usepackage{booktabs}
\usepackage{amsmath}
\usepackage{comment}

\usepackage{csquotes}
\newcommand{\dq}[1]{\enquote{#1}}

\usepackage{array}

\aclfinalcopy 
\def\aclpaperid{790} 

\title{Keyphrase Generation: A Text Summarization Struggle}

\author{Erion \c{C}ano \\
  Institute of Formal and Applied \\
  Linguistics, Charles University, \\
  Prague, Czech Republic \\
  {\tt cano@ufal.mff.cuni.cz} \\\And
  Ond\v{r}ej Bojar \\
  Institute of Formal and Applied \\
  Linguistics, Charles University, \\
  Prague, Czech Republic \\
  {\tt bojar@ufal.mff.cuni.cz} \\}

\date{}

\begin{document}
\maketitle
\begin{abstract}
Authors' keyphrases assigned to scientific articles are essential for recognizing content and topic aspects. Most of the proposed supervised and unsupervised methods for keyphrase generation are unable to produce terms that are valuable but do not appear in the text. In this paper, we explore the possibility of considering the keyphrase string as an abstractive summary of the title and the abstract. First, we collect, process and release a large dataset of scientific paper metadata that contains 2.2 million records. Then we experiment with popular text summarization neural architectures. Despite using advanced deep learning models, large quantities of data and many days of computation, our systematic evaluation on four test datasets reveals that the explored text summarization methods could not produce better keyphrases than the simpler unsupervised methods, or the existing supervised ones.  
\end{abstract}
%
%
\section{Introduction}
\label{sec:intro}
A valuable concept for searching and categorizing scientific papers in digital libraries is the \emph{keyphrase} (we use \emph{keyphrase} and \emph{keyword} interchangeably), a short set of one or few words that represent concepts. Scientific articles are commonly annotated with keyphrases based on taxonomies of concepts and the authors' judgment. Finding keyphrases that best describe the contents of a document is thus essential and rewarding.

Most of the proposed keyphrase extraction solutions tend to be unsupervised \cite{P17-1102, Nguyen:2010:WKE:1859664.1859699, Rose2010, I13-1062, 10.1007/978-3-319-76941-7_80} and generate terms by selecting the most appropriate candidates, ranking the candidates based on several features and finally returning the top $N$. 
Another way is to utilize datasets of texts and keywords for training supervised models with linguistic or other features to predict if candidates are keywords or not \cite{Witten:1999:KPA:313238.313437, Turney:2000:LAK:593957.593993, medelyan2009human, hulth2003improved}.

All above methods propose $N$ keyphrases for each article which are joined together with \dq{,} (or other separator like \dq{;}) to form the \emph{keyphrase string} of that article. They suffer from various problems or discrepancies.  
First, they are unable to find an optimal value for $N$ and require it as a preset parameter. Furthermore, semantic and syntactic properties of article phrases are analyzed separately. The meaning of paragraphs, sections or entire document is thus missed. Lastly, only phrases that do appear in the article are returned. 
\citet{P17-1054} recently proposed a deep supervised keyphrase generation solution trained on a big dataset. It successfully solves the last two problems above, but not the first one.  

Motivated by recent advances in neural machine translation and abstractive text summarization \cite{NIPS2017_7181, DBLP:conf/acl/FosterVUMKFJSWB18, D15-1044, P17-1099}, in this paper, we explore the possibility of considering keyphrase generation as an abstractive text summarization task. Instead of generating keywords one by one and linking them to form the keyphrase string, we consider the later as an abstractive summary of the concatenated paper title and abstract. Different recently-proposed text summarization architectures are tried on four test datasets of article keyphrases \cite{DBLP:journals/corr/abs-1708-02043, D15-1044, P17-1099}. We trained them with a newly created dataset of 2.2 million article titles, abstracts and keyphrase strings that we processed and released.\footnote{\url{http://hdl.handle.net/11234/1-2943}}

The selected text summarization models are compared with popular unsupervised and supervised methods using ROUGE \cite{Lin:2004} and full-match F$_1$ metrics. 
The results show that though trained with large data quantities for many days, the tried text summarization methods could not produce better keywords than the existing supervised or deep supervised predictive models. 
In our opinion, a possible explanation for this is the fact that the title and the abstract may not carry sufficient topical information about the article, even when joined together. In contrast, when assigning keywords annotations of their paper, authors are highly influenced by the topic aspects of it.

This paper carries several contributions, despite the fact that no progressive result scores were reached. It is the first work that considers keyphrase generation as an abstractive text summarization task. We produced a large dataset of article titles, abstracts, and keywords that can be used for keyword generation, text summarization or similar purposes. Finally, we evaluated the performance of different neural network architectures on summarization of article keyword strings, comparing them with popular unsupervised methods.
\section{Scientific Paper Datasets}  
\label{sec:data}
Because of the open source and open data initiatives, many public datasets from various domains can be found online \cite{7325106}. Among the several collections of scientific articles, some of them have gained considerable popularity in research literature.  
In \citet{P17-1054}, we found a recent and big collection of 20K paper abstracts and keyphrases. These metadata belong to articles of computer science from ACM Digital Library, ScienceDirect, and Web of Science.  
In \citet{hulth2003improved}, we found a collection of 2000 (1500 for train/val and 500 for testing) abstracts in English, together with titles and authors' keywords. The corresponding articles were published from 1998 to 2002 and belong to the discipline of \emph{Information Technology}. Furthermore, \citet{krap2010} released a dataset of 2000 (1600 for train/val and 400 for testing) full articles published by ACM from 2003 to 2005 in Computer Science domain. 
More information about similar keyphrase data collections or other available resources can be found in \citet{Hasan+Ng:14a} and in online \href{https://github.com/LIAAD/KeywordExtractor-Datasets}{repositories}.

Regarding text summarization, some of the most popular datasets are: DUC-2004
\footnote{\url{https://duc.nist.gov/duc2004/}} 
mainly used for testing, English Gigaword \cite{Napoles:2012:AG:2391200.2391218}, CNN/Daily~Mail described in Section~4.3 of \cite{K16-1028} and Newsroom, a heterogeneous bundle of news articles described in \citet{N18-1065}. These datasets are frequently used for the task of predicting titles from abstracts or short stories. However, no keyphrases are provided; they do not serve to our purpose.
ArnetMiner is a recent attempt to crawl scientific paper data from academic networks \cite{tang2008arnetminer}. The system extracts profiles of researchers from digital resources and integrates their data in a common network. A spin-off is the Open Academic Graph (OAG) data collection \cite{Sinha:2015:OMA:2740908.2742839}.
\begin{table}[!t]
\centering
\small
\begin{tabular}{|l|c|c|c|c|}
\hline
\bf Attribute & \bf Train & \bf Val & \bf Test & \bf Fullset \\ [0.1ex] 
\hline
Records & 2M & 100K & 100K & 2.2M \\ [0.07ex]
Keyphrases & 12M & 575K & 870K & 13.4M \\ [0.07ex]
Title tokens & 24M & 1.3M & 1.6M & 27M \\ [0.07ex]
Abstract tokens & 441M & 21M & 37M & 499M \\ [0.07ex]
\hline
Av. Keyphrase & 6 & 5.8 & 8.7 & 6.1 \\ [0.07ex]
Av. Title & 12.1 & 12.8 & 15.9 & 12.3 \\ [0.07ex]
Av. Abstract & 220 & 211 & 372 & 227 \\ [0.07ex]
\hline
\end{tabular}
\caption{Statistics of OAGK dataset}
\label{table:dataStats}
\end{table}

To produce a usable collection for our purpose, we started from OAG. We extracted \emph{title}, \emph{abstract} and \emph{keywords}.
The list of keywords was transformed into a comma-separated string and  
a language identifier was used to remove records that were not in English.
Abstracts and titles were lowercased, and Stanford CoreNLP tokenizer was used for tokenizing. 
Short records of fewer than 20 tokens in the abstract, 2 tokens in the title and 2 tokens in the keywords were removed. 
For the test portion, we selected documents of at least 27, 3 and 2 tokens in each field. Data preprocessing stopped here for the release version (no symbol filtering), given that many researchers want to filter text in their own way. This new dataset named OAGK can be used for both text summarization (predicting title from abstract) and keyphrase extraction (unsupervised, supervised or deep supervised) tasks. Some rounded measures about each set of released data are presented in Table~\ref{table:dataStats}. 
\section{Keyphrase Extraction Strategies}
\label{sec:keyphraseMethods}
\subsection{Unsupervised and Supervised Methods}
\label{ssec:Unsupervised}
\textsc{TopicRank} is an extractive method that creates topic clusters using the graph of terms and phrases \cite{I13-1062}. Obtained topics are then ranked according to their importance in the document. Finally, keyphrases are extracted by picking one candidate from each of the most important topics. 
A more recent, unsupervised and feature-based method for keyphrase extraction is \textsc{Yake!} \cite{10.1007/978-3-319-76941-7_80}. It heuristically combines features like \emph{casing}, \emph{word position} or \emph{word frequency} to generate an aggregate score for each phrase and uses it to select the best candidates.  

One of the first supervised methods is \textsc{Kea} described by \citet{Witten:1999:KPA:313238.313437}. It extracts those candidate phrases from the document that have good chances to be keywords. Several features like \emph{TF-IDF} are computed for each candidate phrase during training. In the end, Na\"ive Bayes algorithm is used to decide if a candidate is a keyword or not (binary classification). 
An improvement and generalization of \textsc{Kea} is \textsc{Maui} \cite{medelyan2009human}. Additional features are computed, and bagged decision trees are used instead of Na\"ive Bayes. The author reports significant performance improvements in precision, recall and F$_1$ scores.  

The above keyphrase extraction methods and others like \citet{P17-1102} or \citet{Nguyen:2010:WKE:1859664.1859699} 
reveal various problems.
First, they are not able to find an optimal value for $N$ (number of keywords to generate for an article) based on article contents and require it as a preset parameter.  
Second, the semantic and syntactic properties of article phrases (considered as candidate keywords) are analyzed separately. The meaning of longer text units like paragraphs or entire abstract/paper is missed. 
Third, only phrases that do appear in the paper are returned. In practice, authors do often assign words that are not part of their article. 

\citet{P17-1054} overcome the second and third problem using an encoder-decoder model (\textsc{CopyRnn}) with a bidirectional Gated Recurrent Unit (GRU) and a forward GRU with attention. They train it on a datasets of hundred thousands of samples, consisting of abstract-keyword (one keyword only) pairs. The model is entirely data-driven and can produce terms that may not appear in the document.
It still produces one keyword at a time, requiring $N$ (first problem) as parameter to create the full keyphrase string.  
\begin{table*}[ht]
\small 
\centering      
\setlength\tabcolsep{7pt}  
\begin{tabular}
{| l | r r | r r | r r | r r |}
\hline
& \multicolumn{8}{|c|}{\textbf{~~Hulth (500)} \qquad \textbf{Krapivin (400)} \qquad \textbf{Meng (20K)} \qquad~
\textbf{OAGK (100K)}} \\ [0.12ex] 
\textbf{Method} & \textbf{F$_1$@5} & \textbf{F$_1$@7} & \textbf{F$_1$@5} & \textbf{F$_1$@7} &
\textbf{F$_1$@5} & \textbf{F$_1$@7} & \textbf{F$_1$@5} & \textbf{F$_1$@7} \\ [0.12ex] 
\hline
\textsc{Yake!} & 19.35 & 21.47 & 17.98 & 17.4 & 17.11 & 15.19 & 15.24 & 14.57 \\ [0.07ex]
\textsc{TopicRank} & 16.5 & 20.44 & 6.93 & 6.92 & 11.93 & 11.72 & 11.9 & 12.08 \\ [0.07ex]
\textsc{Maui} & 20.11 & 20.56 & 23.17 & 23.04 & 22.3 & 19.63 & 19.58 & 18.42 \\ [0.07ex]
\textsc{CopyRnn} & {\bf 29.2} & {\bf 33.6} & {\bf 30.2} & {\bf 25.2} & {\bf 32.8} & {\bf 25.5} & {\bf 33.06} & {\bf 31.92} \\ [0.07ex]
\hline
\textsc{Merge} & 6.85 & 6.86 & 4.92 & 4.93 & 8.75 & 8.76 & 11.12 & 13.39 \\ [0.07ex]
\textsc{Inject} & 6.09 & 6.08 & 4.1 & 4.11 & 8.09 & 8.09 & 9.61 & 11.22 \\ [0.07ex]
\textsc{Abs} & 14.75 & 14.82 & 10.24 & 10.29 & 12.17 & 12.09 & 14.54 & 14.57 \\ [0.07ex]
\textsc{PointCov} & 22.19 & 21.55 & 19.87 & 20.03 & 20.45 & 20.89 & 22.72 & 21.49 \\ [0.07ex]
\hline
\end{tabular} 
\caption{Full-match scores of predicted keyphrases by various methods} 
\label{table:exactMatch}
\end{table*}
\begin{table*}[ht]
\small 
\centering      
\setlength\tabcolsep{7pt}  
\begin{tabular}
{| l | r r | r r | r r | r r |}
\hline
& \multicolumn{8}{|c|}{\textbf{~~Hulth (500)} \qquad \textbf{Krapivin (400)} \qquad \textbf{Meng (20K)} \qquad~
\textbf{OAGK (100K)}} \\ [0.12ex] 
\textbf{Method} & $\boldsymbol{R_1F_1}$ & $\boldsymbol{R_LF_1}$ & 
$\boldsymbol{R_1F_1}$ & $\boldsymbol{R_LF_1}$ & $\boldsymbol{R_1F_1}$ & 
$\boldsymbol{R_LF_1}$ & $\boldsymbol{R_1F_1}$ & $\boldsymbol{R_LF_1}$ \\ [0.12ex] 
\hline
\textsc{Yake!} & 37.48 & 24.83 & 26.19 & 18.57 & 26.47 & 17.36 & 20.38 & 14.54 \\ [0.07ex]
\textsc{TopicRank} & 32.0 & 20.36 & 14.08 & 11.47 & 21.68 & 15.94 & 17.46 & 13.28 \\ [0.07ex]
\textsc{Maui} & 36.88 & 27.16 & 28.29 & 23.74 & 34.33 & 28.12 & 32.16 & 25.09 \\ [0.07ex]
\textsc{CopyRnn} & {\bf 44.58} & {\bf 35.24} & {\bf 39.73} & {\bf 30.29} & {\bf 42.93} &  34.62 & {\bf 43.54} & {\bf 36.09} \\ [0.07ex]
\hline 
\textsc{Merge} & 15.19 & 9.45 & 9.66 & 7.14 & 16.53 & 12.31 & 17.3 & 14.43 \\ [0.07ex]
\textsc{Inject} & 14.15 & 8.81 & 9.58 & 6.79 & 15.6 & 11.21 & 14.3 & 11.08 \\ [0.07ex]
\textsc{Abs} & 27.54 & 19.48 & 25.59 & 18.2 & 28.31 & 22.16 & 29.05 & 25.77 \\ [0.07ex]
\textsc{PointCov} & 37.16 & 33.69 & 35.81 & 29.52 & 38.47 & {\bf 35.06} & 38.66 & 34.04 \\ [0.07ex]
\hline 
\end{tabular}
\caption{Rouge scores of predicted keyphrases by various methods}  
\label{table:rougeScore}
\end{table*}
\subsection{Text Summarization Methods}
\label{ssec:textsumm}
To overcome the three problems mentioned in Section~\ref{ssec:Unsupervised}, we explore abstractive text summarization models proposed in the literature, trained with article abstracts and titles as sources and keyword strings as targets. They are expected to learn and paraphrase over entire source text and produce a summary in the form of a keyphrase string with no need for extra parameters. They should also introduce new words that do not appear in the abstract.  
Two simple encoder-decoder variants based on LSTMs are described in Figure~3 of \citet{DBLP:journals/corr/abs-1708-02043}. \textsc{Merge} (Figure~3.a) encodes input and the current summary independently 
and merges them in a joint representation which is later decoded to predict the next summary token. \textsc{Inject} model (Figure~3.b) on the other hand injects the source document context representation to the encoding part of the current summary before the decoding operation is performed.  

\textsc{Abs} 
is presented in Figure~3.a of \citet{D15-1044}. The encoder (Figure~3.b) takes in the input text and a learned soft alignment between the input and the summary, producing the context vector. This soft alignment is the attention mechanism \cite{DBLP:journals/corr/BahdanauCB14}. To generate the summary words, \citeauthor{D15-1044} apply a beam-search decoder with a window of $K$ candidate words in each position of the summary. 

Pointer-Generator network (\textsc{PointCov}) depicted in Figure~3 of \citet{P17-1099} is similar to \textsc{Abs}. It is composed of an attention-based encoder that produces the context vector. The decoder is extended with a pointer-generator model that computes 
a probability $p_{gen}$ from the context vector, the decoder states, and the decoder output. 
That probability is used as a switch to decide if the next word is to be generated or copied from the input. This model is thus a compromise between abstractive and extractive (copying words from input) models. Another extension is the coverage mechanism 
for avoiding word repetitions in the summary, a common problem of encoder-decoder summarizers \cite{P16-1008}. 
\section{Results}
We performed experiments with the unsupervised and supervised methods of Section~\ref{sec:keyphraseMethods} on the first three datasets of Section~\ref{sec:data} and 
on OAGK. All supervised methods were trained with the 2M records of OAGK train part. An exception was \textsc{Maui} which could be trained on 25K records at most (memory limitation). 
In addition to the processing steps of Section~\ref{sec:data}, we further replaced digit symbols with \# and
limited source and target text lengths to 270 and 21 tokens, respectively. Vocabulary size was also limited to the 90K most frequent words.

The few parameters of the unsupervised methods (length and windows of candidate keyphrases for \textsc{Yake!}, ranking strategy for \textsc{TopicRank}) were tuned using the validation part of each dataset. 
For the evaluation, we used F$_1$ score of full matches between predicted and authors' keywords. 
Given that the average number of keywords in the data is about 6, we computed F$_1$ scores on top 5 and top 7 returned keywords (\textbf{F$_1$@5, F$_1$@7}).

Before each comparison, both sets of terms were stemmed with Porter Stemmer and duplicates were removed. In the case of summarization models, keyphrases were extracted from their comma-separated summaries. 
We also computed ROUGE-1 and ROUGE-L F$_1$ scores ($\boldsymbol{R_1F_1}$, $\boldsymbol{R_LF_1}$) that are suitable for evaluating short summaries \cite{Lin:2004}. 
The keywords obtained from the unsupervised methods were linked together to form the keyphrase string (assumed summary). This was later compared with the original keyphrase string of the authors.  

Full-match results on each dataset are reported in Table~\ref{table:exactMatch}. From the unsupervised models, we see that \textsc{Yake!} is consistently better than \textsc{TopicRank}. The next two supervised models perform even better, with \textsc{CopyRnn} being discretely superior than \textsc{Maui}.

Results of the four summarization models seem disappointing. \textsc{Merge} and \textsc{Inject} are the worst on every dataset, with highest score 13.39~\%. 
Various predictions of these models are empty or very short, and some others contain long word repetitions which are discarded during evaluation. As a result, there are usually fewer than five predicted keyphrases. This explains why \textbf{F$_1$@5} and \textbf{F$_1$@7} scores are very close to each other. 

\textsc{Abs} works slightly better reaching scores from 10.24 to 14.75~\%. \textsc{PointCov} is the best of the text summarizers producing keyphrase predictions that are usually clean and concise with few repetitions. This is probably the merit of the coverage mechanism. There is still a considerable gap between \textsc{PointCov} and \textsc{CopyRnn}.   
Rouge-1 and Rouge-L F$_1$ scores are reported in Table~\ref{table:rougeScore}. 
\textsc{CopyRnn} is still the best but \textsc{PointCov} is close. \textsc{Abs} scores are also comparable to those of \textsc{Maui} and \textsc{Yake!}. \textsc{TopicRank}, \textsc{Merge} and \textsc{Inject} are again the worst. 

Regarding the test datasets, the highest result scores are achieved on Hulth and the lowest on Krapivin. We checked some samples of the later and observed that each of them contains separation tags (e.g., --T, --A, --B, Figure etc.) for indicating different parts of text in the original paper. A more intelligent text cleaning step may be required on those data.  
\section{Discussion}
\label{sec:Discussion}
The results show that the tried text summarization models perform poorly on full-match keyword  predictions. Their higher ROUGE scores further indicate that the problem is not entirely in the summarization process. 
Observing a few samples, we found differences between the two evaluation strategies. For example, suppose we have the predicted keyword \emph{\dq{intelligent system}} compared against authors' keyword \emph{\dq{system design}}. Full-match evaluation adds nothing to \textbf{F$_1$@5} and \textbf{F$_1$@7} scores. However, in the case of ROUGE evaluation, the prediction is partially right and a certain value is added to $\boldsymbol{R_1F_1}$ score. In follow up works, one solution to this discrepancy could be to try partial-match comparison scores like overlap coefficients.  

Another detail that has some negative effect in full-match scores is keyword separation. The predicted string: 
\begin{center}
\emph{\dq{health care,,,,immune system; human -; metabolism, immunity,,,,}}  \\
\end{center}
produces [\dq{health care}, \dq{immune system}, \dq{human}, \dq{metabolism}, \dq{immunity}] as the list of keywords after removing the extra separators. Instead, we expected [\dq{health care}, \dq{immune system}, \dq{human metabolism}, \dq{immunity}]. This again penalizes full-match scores but not $\boldsymbol{R_1F_1}$ score. A more intelligent keyword separation mechanism could thus help for higher full-match result scores. 

A third reason could be the fact that we used the title and abstract of papers only. This is actually what most researchers do, as it is hard to find high quantities of article full texts for free. Article body is usually restricted. Abstractive summarization methods could still benefit from longer source texts.
Using default hyperparameters for the models may have also influenced the results. Some parameter tuning could be beneficial, though.

The main reason could be even more fundamental. We trained abstractive summarization models on abstracts and titles with authors' keyphrases considered as golden ones. There might be two issues here. First, when setting their keywords, authors mostly consider the topical aspects of their work rather than paraphrasing over the contents. Abstracts and titles we used may not carry enough topical information about the article, even when joined together. Second, considering authors' keywords as golden ones may not be reasonable. One solution is to employ human experts and ask them to annotate each article based on what they read. This is however prohibitive when hundred thousands of samples are required.
Extensive experiments on this issue may provide different facts and change the picture. For the moment, a safe way to go seems developing deep supervised generative models like the one of \citet{P17-1054} that predict one keyphrase at each step independently. 
\section{Conclusions}
\label{sec:conclusions}
In this paper, we experimented with various unsupervised, supervised, deep supervised and abstractive text summarization models for predicting keyphrases of scientific articles. To the best of our knowledge, this is the first attempt that explores the possibility of conceiving article string of keywords as an abstractive summary of title and abstract. We collected and produced a large dataset of 2.2 million abstracts, titles and keyphrase strings from scientific papers available online. It can be used for future text summarization and keyphrase generation experiments. Systematic evaluation on four test datasets shows that the used summarization models could not produce better keywords than the supervised predictive models. Extensive experiments with more advanced summarizaiton methods and better parameter optimization may still reveal a different view of the situation. 
\section*{Acknowledgments}
This research work was supported by the project No. CZ.02.2.69/0.0/0.0/16\_027/0008495 (International Mobility of Researchers at Charles University) of the Operational Programme Research, Development and Education, grant 19-26934X (NEUREM3) of the Czech Science Foundation and H2020-ICT-2018-2-825460 (ELITR) of the EU. %
\bibliography{naaclhlt2019}
\bibliographystyle{acl_natbib}
\end{document}